\documentclass[11pt,a4paper]{article}
\usepackage[hyperref]{AACL-IJCNLP2020}
\usepackage{times}
\usepackage{latexsym}

\usepackage{microtype}

\aclfinalcopy 

\title{Towards Code-switched Classification Exploiting Constituent Language Resources}

\author{
    Tanvi Dadu \thanks{\hspace{2 mm} Both authors contributed equally to the work.} \and Kartikey Pant \footnotemark[1] \\\
    Netaji Subhas Institute of Technology, New Delhi, India \\
    International Institute of Information Technology, Hyderabad, India \\
    {\tt tanvid.co.16@nsit.net.in} \\
    {\tt kartikey.pant@research.iiit.ac.in} \\
}

\date{}
\usepackage{graphicx}
\usepackage{multirow}
\usepackage[normalem]{ulem}
\useunder{\uline}{\ul}{}
\begin{document}
\maketitle
\begin{abstract}
Code-switching is a commonly observed communicative phenomenon denoting a shift from one language to another within the same speech exchange. The analysis of code-switched data often becomes an assiduous task, owing to the limited availability of data. We propose converting code-switched data into its constituent high resource languages for exploiting both monolingual and cross-lingual settings in this work. This conversion allows us to utilize the higher resource availability for its constituent languages for multiple downstream tasks.

We perform experiments for two downstream tasks, sarcasm detection and hate speech detection, in the English-Hindi code-switched setting. These experiments show an increase in $22\%$ and $42.5\%$ in \textit{F1-score} for sarcasm detection and hate speech detection, respectively, compared to the state-of-the-art.
\end{abstract}

\section{Introduction}

Code-switching is the juxtaposition of speech belonging to two different grammatical systems or subsystems within the same speech exchange. In other words, it denotes a shift from one language to another within the same utterance but not limited to beyond mere insertion of borrowed words, fillers, and phrases. It often involves morphological and grammatical mixing\cite{SurveyPaper}. Code-switching of speech is a common phenomenon in the multilingual communities \cite{HandBookOfContact}, especially among peers who have similar fluency in multiple languages. Code-switching among languages like Spanish-English, Hindi-English, Tamil-English is often seen to be used by bilingual speakers of these languages.

Code-switching is likely to be found in informal settings, including speech and social media, and semi-formal such as navigation instructions. This phenomenon leads us to acknowledge code-switching as a legitimate communication form deserving careful linguistic analysis and processing. Processing code-switched communication would enhance user experience in various industrial settings, including advertising, healthcare, education, and entertainment \cite{SurveyPaper}. 

Access to code-switched data is challenging and limited. This phenomenon makes the analysis and information extraction from code-switched languages a less explored and challenging task. Most code-switching studies focus on pairs of a high resource and a lower resource language, often making these studies data starved. Also, code-switched language is mostly used for informal and semi-formal settings, making it less archived, resulting in its limited availability in terms of context and volume.

Traditionally, most works in academia have limited themselves to language-independent methodologies like SVM, FastText, and CNN \cite{SurveyPaper} for modeling code-switched languages. This can be attributed to the shortage of data in code-switched settings, limiting the exploitation of predictive performance achieved via transfer learning using pretrained contextualized word embeddings. 

We propose a generic methodology for modeling Hindi-English code-switched data revolving around its translation and transliteration. We propose two approaches that exploit a) Transliteration to convert this low-resource task into a high-resource task using cross-lingual learning, and b) Translation to utilize monolingual resources of each of the constituent languages. The importance of this methodology is that we enable the exploitation of pretrained transformer-based embeddings that are available in the monolingual or cross-lingual setting but not code-mixed setting due to the scarcity of data for pretraining. We then benchmark both the approaches for the tasks of hate speech detection and sarcasm detection using the current state-of-the-art and language-independent supervision learning as the baseline. We further provide insights into the proposed approaches and compare them analytically.

\section{Related Works}

The previous works on sarcasm and humor detection contain a myriad of methods employed over the years on multiple datasets, including statistical and N-gram analysis on spoken conversations from an American sitcom \cite{purandare2006}, Word2Vec combined with KNN Human Centric Features on the \textit{Pun of the Day} and \textit{16000 One-Liner datasets} \cite{Yang2015}, and Convolutional Neural Networks on datasets with distinct joke types in both English and Chinese \cite{Chen2018}. Recently, transformer-based architectures have been used to detect humor on multiple datasets collected from Reddit, Kaggle, and Pun of the Day Dataset \cite{Weller2019}.

The prior works exploring hate speech detection employ several machine learning-based methods like linguistic features with character n-grams \cite{Waseem2016}, SVM with a set of features that includes n-grams, skip-grams, and clustering-based word representations \cite{Malmasi2018}. Neural networks like LSTM \cite{Badjatiya2017}, FastText \cite{Badjatiya2017} and CNN \cite{Gamback2017} have also been used for hate speech detection. Recently, transformer-based architectures using BERT, ALBERT, and RoBERTa have been used to detect hate speech ~\cite{Mozafari2019, Wiedemann2020}.

There has been considerable work in code-switching, as is well documented by \citet{SurveyPaper}. The use of code-switched data has been explored for the tasks of Part-of-Speech tagging \cite{Vyas2014}, Named Entity Recognition \cite{Aguilar2018}, Dependency Parsing \cite{Partanen2018}, and Text Classification \cite{Swami2018,Bohra2018}.

\citet{Swami2018} introduced an English-Hindi code-switching dataset for sarcasm detection in social media text. Their dataset contains tweets annotated as sarcastic and non-sarcastic in the form of their constituent language-identified tokens. They further provide a feature-based model as a baseline. We use their dataset for our sarcasm detection experiments.

\citet{Bohra2018} introduced a dataset of English-Hindi social media text, demonstrating code-switching for the task of hate speech detection. They also conduct benchmarking experiments using a feature-based model. Similar to \citet{Swami2018}, their benchmarks only contain accuracies for their baseline systems and do not take F1-scores into account. We use their dataset for our hate speech detection experiments.

There have been attempts to exploit monolingual datasets for enhancing downstream tasks in a code-switched setting. \citet{Solorio2008} explored the use of monolingual taggers for Spanish and English for Part-of-Speech Tagging in an English-Spanish code-switched dataset. They further show that their best results were obtained by using the output of monolingual POS taggers as input features. \citet{Gupta2018} explore using monolingual and bilingual resources for the task of Question Answering in an English-Hindi code-switched dataset.

For text classification involving code-switched text, traditional methodologies like SVM, FastText, CNN has been used \cite{SurveyPaper}. However, most works are unable to exploit the recent advances in pretrained models, including RoBERTa and XLM-RoBERTa \cite{2020roberta, XLMRoBERTa} due to data scarcity and lack of normalization in the code-switched setting. To overcome this, \citet{srivastava2020hcms} attempt to use multilingual BERT for code-switched sentiment analysis of social media text using a transliteration based methodology. However, they failed to beat the traditional baselines, including fasttext.

\begin{figure*}[!h]
        \centering
        \includegraphics[width=16cm]{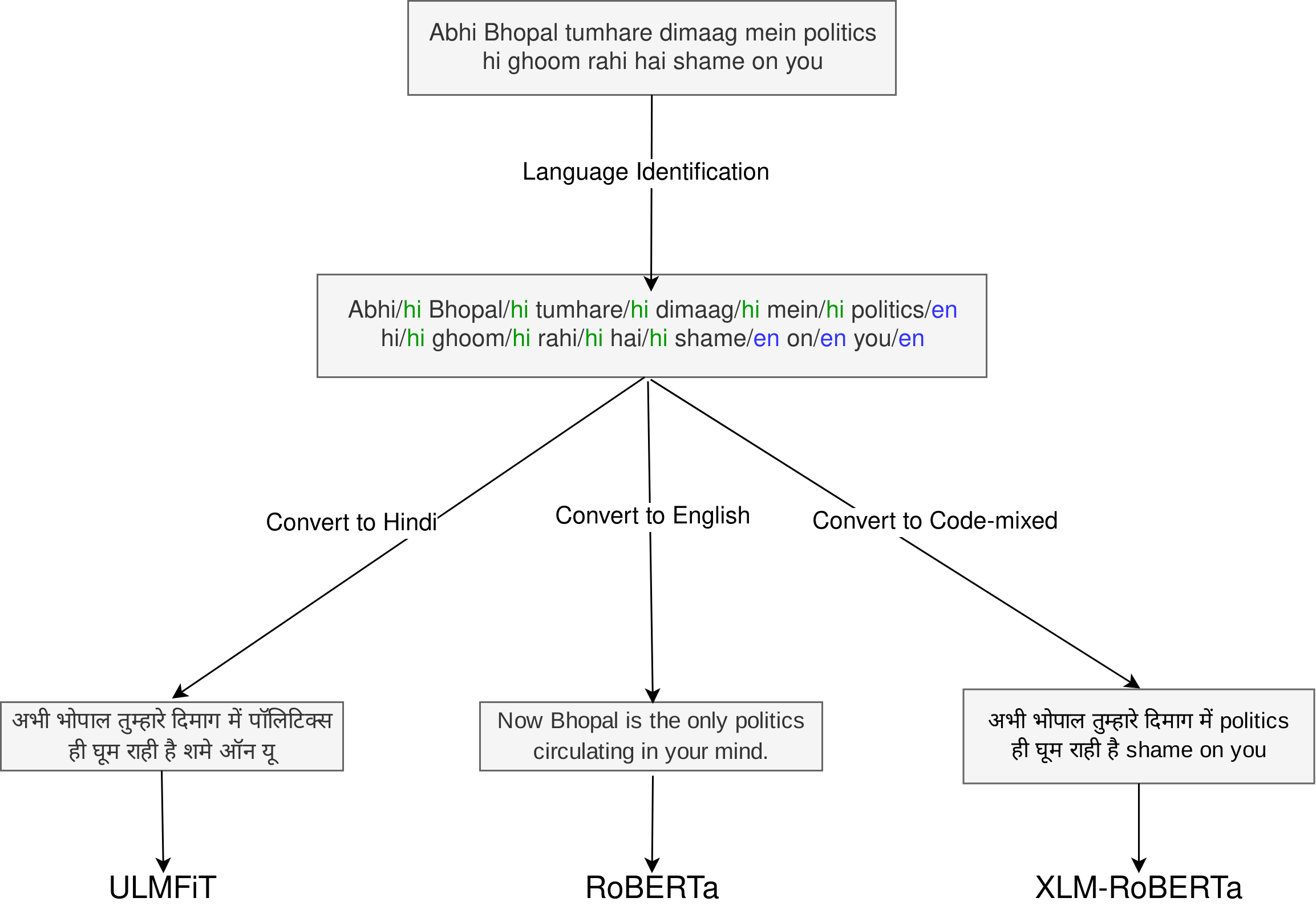}
        \caption{An example sentence demonstrating the proposed approach.}
        \label{fig:example-architecture}
\end{figure*}
 
\section{Approach}

This section outlines our approach for analyzing code-switched language by converting it into its respective high resource languages and fine-tuning monolingual and cross-lingual contextual word embeddings. We highlight the data preparation followed by the contextual embedding used for each of the variants. \autoref{fig:example-architecture} illustrates our proposed approach through an example code-switched sentence from one of the datasets.

\subsection{High Resource Language A - Hindi}

\subsubsection{Data Preparation}

This section outlines our approach for analyzing code-switched language by converting it into its respective high resource languages and fine-tuning monolingual and cross-lingual contextual word embeddings. We highlight the data preparation followed by the contextual embedding used for each of the variants. 

This section presents the steps performed for converting the code-switched data into one of the high resource languages, Hindi:
\begin{enumerate}
\item \textbf{Language Identification}: Each token was identified as either containing Hindi, English, or neither of them. 
\item \textbf{Transliteration into Devanagari}: Each language-marked token, containing Hindi and English words, was transliterated from the Latin script to the Devanagari script. We use Microsoft Translate Text \footnote{\href{https://www.microsoft.com/en-us/translator/business/translator-api/}{{https://www.microsoft.com/en-us/translator/business/translator-api/}}} service's Transliteration API for the transliteration process.   
\end{enumerate}

Language Identification for the Hate Speech Detection task was done using a supervised FastText model trained on the Sarcasm Detection corpus \cite{Swami2018} containing tokens with their respective languages- English, Hindi, and Other (includes Hashtags, Mentions, and Punctuation marks).

\begin{table}[h]
\centering
\resizebox{0.48\textwidth}{!}{%
\begin{tabular}{|l|r|}
\hline
{ \textbf{Dataset}} & \multicolumn{1}{l|}{{ \textbf{Percentage of Hindi Tokens}}} \\ \hline
{ \textbf{Sarcasm Detection}} & { 88.259\%} \\ \hline
{ \textbf{Hate Speech}} & { 86.437\%} \\ \hline
\end{tabular}%
}
\caption{Percentage of Hindi Tokens in comparison to Language-marked Tokens.}
\label{tab:hindi-proportion}
\end{table}

\autoref{tab:hindi-proportion} shows that the majority of the words present in the dataset are Hindi, allowing us to assume that code-switched language contains Hindi and English words borrowed into Hindi. This phenomenon allows us to take advantage of converting all words from Latin to Devanagari via transliteration.

\subsubsection{Classification Model}

For the classification of the text translated to Hindi, we use Universal Language Model Fine-tuning for Text Classification, ULMFiT \cite{Howard2018}. We use a pretrained language model trained on $172K$ Wikipedia articles in Hindi, attaining a perplexity of $34.06$ \footnote{\href{https://github.com/goru001/nlp-for-hindi}{https://github.com/goru001/nlp-for-hindi}}. We fine-tune our language model using discriminative fine-tuning, using different learning rates for different layers of the language model and slanted triangular learning rates. We then fine-tune the final model with the text translated in Hindi, after augmenting the trained language model with two additional linear blocks for classification. These blocks use batch normalization and dropout, with ReLU activations for the intermediate layer and a softmax activation for the last layer.

\subsection{High Resource Language B - English}

\subsubsection{Data Preparation}
This section outlines the conversion process of the code-switched data into the other high resource language, English:

\begin{enumerate}
\item \textbf{Language Identification}: Similar to the previous subsection, each token was identified to be either containing Hindi, English, or neither of them. 
\item \textbf{Transliteration into Devanagari}: We used the same model as in the previous subsection for transliterating each language-marked token from the Latin script to the Devanagari script. 
\item \textbf{Translation into English}: Each full sentence from the transliterated text was translated into English using the Translate API of the Microsoft Translate Text service.
\end{enumerate}

\subsubsection{Classification Model}
For classification of the generated English text, we use a strong model for monolingual English classification, $RoBERTa_{Large}$. RoBERTa \cite{2020roberta} is a contextualized word representation model, pre-trained using a bidirectional Transformer-based encoder. It is trained using significantly larger data with carefully tuned hyperparameters and performs competitively in benchmarking texts like GLUE \cite{GLUE} for text classification and SQuAD \cite{SQUAD} for question answering.  Since the model is pretrained on a large generic dataset, it can be fine-tuned for a specific task in a simpler yet efficient manner without making significant changes in its architecture. We fine-tune its large variant, $RoBERTa_{Large}$, for both the tasks.

\subsection{Cross-lingual Classification}

\subsubsection{Data Preparation}

Each language-marked token needs to have its correct script to utilize cross-lingual word representations.  We, therefore, employ the following steps to prepare the dataset for cross-lingual classification:

\begin{enumerate}
    \item \textbf{Language Identification}: Similar to the previous subsections, each token was identified to be either containing Hindi, English, or neither of them. 
    \item \textbf{Transliteration of Hindi words into Devanagari}: We used the same model as in the previous subsections for transliterating each token marked to be Hindi from the Latin script to the Devanagari script.
\end{enumerate}

\subsubsection{Classification Model}
For cross-lingual classification, we use XLM-RoBERTa, a state-of-the-art contextual word representation released by Facebook  \cite{XLMRoBERTa}. XLM-R is a cross-lingual unsupervised contextual word representation pretrained on a 100-language sized dataset. We exploit its advantage of massive cross-lingual transfer learning by fine-tuning it for both the tasks. Due to its high scalability and ease of use in fine-tuning, the model enables us to achieve competitive predictive performance while using minimum training resources. We use the large variant, XLM-R, for both the tasks.

\section{Experimental Results}

\subsection{Downstream Tasks}
This section introduces the datasets used for the downstream tasks of Sarcasm Detection and Hate Speech Detection.

For the task of sarcasm detection, we use English-Hindi code-switched dataset released by \citet{Swami2018}. The dataset contains $5,250$ tweets sampled from Twitter with sarcastic or non-sarcastic tags. It also provides language tags (English/Hindi/Others) for all the tokens present in tweets. The dataset has $0.790$ Cohen's Kappa score for measuring inter-annotator agreement, implying a high quality of the annotation schema. 

For the task of hate speech detection, we use \citet{Bohra2018}'s dataset containing tweets showing English-Hindi code-switching. From their released dataset, we were able to get $4,575$ tweets along with their binary label denoting whether the tweet contains hate speech or not. We preprocess the dataset and remove hyperlinks embedded in the tweets after language identification of each token. The authors measure the inter-annotator agreement through Kohen's Kappa score, which is reasonably high, with a value of $0.982$. 

\subsection{Baselines}
We use the following neural architectures as baselines for our experiments:
\begin{enumerate}
    \item FastText \cite{joulin2016tricks}: FastText is a library released by Facebook which uses bag of words and bag of n-grams as features for text classification. It relies on capturing partial information about the local word order efficiently. 
    \item Convolutional Neural Networks (CNN) \cite{kim2014convolutional}:  Convolutional neural networks are multistage trainable neural network architectures developed for classification tasks employing convolution and pooling for extracting features from the text.
\end{enumerate}

\subsection{Experimental Settings}

In this subsection, we outline the experimental settings used for each of the models used in the experiment. We evaluated our model on a held-out test dataset for all experiments, consisting of $10\%$ of the total dataset. For validation purposes, we further split our training dataset using a $90:10$ train-validation split. We evaluate all the models on the following metrics: \textit{F1}, \textit{Precision}, \textit{Recall}, \textit{Accuracy}.

We use FastText's recently open-sourced automatic hyperparameter optimization functionality and run $100$ trials of optimization. We use a two-dense-layered architecture for Convolutional Neural Network with $3$ and $4$ convolution layers for sarcasm and hate speech detection, respectively. We further used a sequence length of $100$ and an embedding size of $300$ with a dropout rate of $0.1$ and $0.2$.

For ULMFiT, we use a pretrained language model trained on $172K$ Wikipedia articles in Hindi. We use the SentencePiece tokenizer for tokenizing the texts. For language modeling, we used a batch size of $16$ and BPTT of $70$. We use AWD-LSTM architecture with a dropout rate of $0.5$ as a classifier.  For $RoBERTa_{Large}$ and XLM-R, we fine-tune with a learning rate of $1*10^{-5}$ for $3$ epochs, each with a maximum sequence length of $50$.

\subsection{Results}

Table \ref{tab:sarcasm-speech-experiment} and Table \ref{tab:hate-speech-experiment} show the experimental results obtained for the sarcasm detection dataset and hate speech dataset, respectively. We observe that our proposed approach of converting code-switched data into respective high resource language outperforms previously used approaches like FastText and CNN significantly. For both sarcasm detection and hate speech detection, preparing code-switched data for cross-lingual classification and using it to fine-tune XLM-R gives the best results. This is reflected in the model obtaining an \textit{F1-score} of $0.850$ for sarcasm detection and $0.724$ for hate speech detection. For both tasks, respectively, it is $22\%$ and $42.5\%$ higher than \textit{CNN}, the best performing baseline.

\begin{table}[ht]
\centering
\resizebox{0.48\textwidth}{!}{%
\begin{tabular}{|l|r|r|r|r|}
\hline
{\textbf{Model}} & {\textbf{Accuracy}} & {\textbf{Precision}} & {\textbf{Recall}} & {\textbf{F1}} \\ \hline
{\textbf{FastText}} & {$76.22\%$} & {0.245} & {0.951} & {0.390} \\
{\textbf{CNN}} & {$95.71\%$} & {\textbf{0.806}} & {0.610} & {0.694} \\ \hline
{\textbf{ULMFiT}} & {$94.85\%$} & {0.800} & {0.733} & {0.765} \\
\textbf{$RoBERTa_{Large}$} & {$95.99\%$} & {0.791} & {0.883} & {0.835} \\
{\textbf{XLM-R}} & {\textbf{$96.37\%$}} & {\textbf{0.806}} & {\textbf{0.900}} & {\textbf{0.850}} \\ \hline
\end{tabular}%
}
\caption{Experimental results for the sarcasm detection task.}
\label{tab:sarcasm-speech-experiment}
\end{table}

\begin{table}[ht]
\centering
\resizebox{0.48\textwidth}{!}{%
\begin{tabular}{|l|r|r|r|r|}
\hline
\multicolumn{1}{|l|}{\textbf{Model}} &
  \multicolumn{1}{l|}{\textbf{Accuracy}} &
  \multicolumn{1}{l|}{\textbf{Precision}} &
  \multicolumn{1}{l|}{\textbf{Recall}} &
  \multicolumn{1}{l|}{\textbf{F1}} \\ \hline
\textbf{FastText}           & 68.78\%                                             & 0.609          & 0.417                        & 0.495          \\
\textbf{CNN}                & {58.08\%} & 0.411          & 0.664                        & 0.508          \\ \hline
\textbf{ULMFiT} & 68.78\%                                             & 0.545          & {0.506} & 0.525          \\
\textbf{$RoBERTa_{Large}$}  & 71.62\%                                             & 0.707          & 0.716                        & 0.711          \\
\textbf{XLM-R}              & \textbf{71.83\%}                                    & \textbf{0.730} & \textbf{0.718}               & \textbf{0.724} \\ \hline
\end{tabular}%
}
\caption{Experimental results for the hate speech detection task.}
\label{tab:hate-speech-experiment}
\end{table}

Further we observe that our proposed approach performs significantly well for all the metrics taken under consideration with stark increase in \textit{F1-score}. In sarcasm detection, it outperforms \textit{F1-score}, \textit{Precision} where as in hate speech detection, it outperforms \textit{F1-score}, \textit{Precision}, \textit{Recall} and \textit{Accuracy} by a significant margin. In sarcasm detection, a slight increase in \textit{Accuracy} can be attributed to the imbalance distribution of classes, which is reinforced by a significant increase in \textit{F1-score}.

The experimental results obtained for both sarcasm and hate speech detection show our approach's effectiveness in leveraging the pretrained contextualized word embeddings for the code-switched settings. It also successfully tackles data shortage when dealing with code-switched data in pretrained settings by converting code-switched language into its constituent languages.

\section{Conclusion}
In this work, we show that converting code-switched data into its constituent high resource language enables us to exploit the performance of the pretrained models for those languages on downstream tasks. We perform experiments on the English-Hindi code-switching pair and benchmark our approach for sarcasm detection and hate speech detection. The experiments show an improvement of up to $42.5\%$ in \textit{F1-score} compared to strong baselines like CNN. Our findings pave the way for further research to utilize monolingual resources for code-switched data for multiple downstream tasks and extend this methodology for other code-switched language pairs. 

\section*{Acknowledgments}
We would like to thank Sai Krishna Rallabandi for reviewing the drafts and the pre-submission mentoring. We would also like to thank the anonymous reviewers for providing critical suggestions.

\bibliography{anthology,aacl-ijcnlp2020}
\bibliographystyle{acl_natbib}

\end{document}